\documentclass[conference]{IEEEtran}
\IEEEoverridecommandlockouts
\usepackage{cite}
\usepackage{amsmath,amssymb,amsfonts}
\usepackage{algorithmic}
\usepackage{graphicx}
\usepackage{textcomp}
\usepackage{xcolor}
\usepackage{float}
\usepackage{tabularx}
\usepackage{caption}
\usepackage{subcaption}

\usepackage{multirow}
\usepackage{tabularx}
\usepackage{longtable}
\usepackage{booktabs}

\def\BibTeX{{\rm B\kern-.05em{\sc i\kern-.025em b}\kern-.08em
    T\kern-.1667em\lower.7ex\hbox{E}\kern-.125emX}}
\begin{document}

\title{Object-Oriented Semantic Mapping for Reliable UAVs Navigation}

\author{\IEEEauthorblockN{Thanh Nguyen Canh\textsuperscript{1,2}, Armagan Elibol\textsuperscript{2}, Nak Young Chong\textsuperscript{2} and $^*$Xiem HoangVan\textsuperscript{1}}
\IEEEauthorblockA{\textsuperscript{1}University of Engineering and Technology, Vietnam National University \\
Hanoi, Vietnam (\{canhthanh, xiemhoang\}@vnu.edu.vn)
\\
\textsuperscript{2}School of Information Science, Japan Advanced Institute of Science and Technology \\
Nomi, Ishikawa 923-1292, Japan (\{aelibol, nakyoung\}@jaist.ac.jp) \\
$^*$\textit{Corresponding authors}}
}
\maketitle~
\begin{abstract}
To autonomously navigate in real-world environments, special in search and rescue operations, Unmanned Aerial Vehicles (UAVs) necessitate comprehensive maps to ensure safety. However, the prevalent metric map often lacks semantic information crucial for holistic scene comprehension. In this paper, we proposed a system to construct a probabilistic metric map enriched with object information extracted from the environment from RGB-D images. Our approach combines a state-of-the-art YOLOv8-based object detection framework at the front end and a 2D SLAM method - CartoGrapher at the back end. To effectively track and position semantic object classes extracted from the front-end interface, we employ the innovative BoT-SORT methodology. A novel association method is introduced to extract the position of objects and then project it with the metric map. Unlike previous research, our approach takes into reliable navigating in the environment with various hollow bottom objects. The output of our system is a probabilistic map, which significantly enhances the map's representation by incorporating object-specific attributes, encompassing class distinctions, accurate positioning, and object heights. A number of experiments have been conducted to evaluate our proposed approach. The results show that the robot can effectively produce augmented semantic maps containing several objects (notably chairs and desks). Furthermore, our system is evaluated within an embedded computer - Jetson Xavier AGX unit to demonstrate the use case in real-world applications.
\end{abstract}

\begin{IEEEkeywords}
Semantic mapping, UAVs, ROS, Metric map.
\end{IEEEkeywords}
\section{Introduction}

One of the primary goals in the realm of intelligent robotics control is to enable robots to comprehend their surroundings to assist us in different activities~\cite{shi2021rgb}. Particularly, the context of indoor search and rescue operations underscores the critical importance of Unmanned Aerial Vehicles (UAVs) in rapidly surveying hazardous environments and furnishing real-time insights to support emergency responders. The efficacy of such missions relies on the UAV's autonomous navigation capability in navigating intricate and cluttered real-world settings while ensuring safety. While the problem of geometrical mapping and localization is traditionally solved through SLAM (Simultaneous Localization and Mapping) seeking a given type of object requires scene knowledge. In addition, autonomous navigation is more challenging in complex environments. Normally, UAVs might encounter diverse obstacles and need to identify victims or hazardous areas, which right away avoid the obstacle. Therefore, UAVs to able to recognize what and where obstacles and navigate through them.

Semantic mapping is integrated within the framework of SLAM, which has emerged as a potential solution to address the challenges of scene understanding in robotics by incorporating the prior or expert information of the environment. Recent research endeavors have predominantly centered around constructing 3D semantic maps utilizing technologies such as depth cameras~\cite{shi2021rgb},~\cite{deng2020semantic}, stereo cameras~\cite{li2018dense}, 3D LiDAR~\cite{chen2019suma}, or fusion sensors~\cite{bersan2018semantic}. However, these approaches often demand substantial computational resources and storage, making them less suitable for UAVs with stringent weight and computation constraints and challenging to use and exploit during UAV navigation. Several kinds of research have also implemented semantic mapping for applications such as room categorization~\cite{setiono2021novel}, and detection dynamic object~\cite{wu20223d}. However, major real-world robots are implemented based on 2D metric maps because of ease of use and low resource consumption. Furthermore, the mapping of hollow bottom objects like tables, desks, and chairs are usually mapped on the map as points or paths leading to wrong navigation decisions. On the other side, recent advancements in deep learning have facilitated the development of robust object detection methods like YOLOv5~\cite{jocher2020ultralytics}, YOLOv7~\cite{wang2023yolov7}, YOLOv8~\cite{Jocher_YOLO_by_Ultralytics_2023}, Fast R-CNN~\cite{ren2015faster}, and Mask R-CNN~\cite{he2017mask}. Hence, our research aims to create an object-oriented semantic mapping for enabling UAVs to perform safe navigation tasks in complex environments.

In this paper, we proposed a system that encompasses a semantic map, forged through the fusion of metric environment structure and semantic object information. We propose an association method that seamlessly integrates object data to create a probabilistic map, thereby enhancing navigation and obstacle avoidance. Remarkably, our system is designed to harness the capabilities of an RGB-D camera, effectively circumventing cost and weight constraints prevalent in UAV applications. Our contributions in this study include:
\begin{itemize}
    \item A semantic SLAM system that combines localization and metric mapping with object tracking to obtain the ability to scene understanding.
    \item A method to extract object information and fusion into to create a probability map for navigation and obstacle avoidance.
    \item The proposed system demonstrated efficiency by being implemented in a Jetson Xavier AGX embedded computer to run in real-time.
\end{itemize}
 The remainder of this paper is organized as follows: Sec.~\ref{sec:method} details the methodology behind our proposed system, emphasizing the integration of object detection, SLAM, and association method. The experiments conducted and results analysis are presented in Sec.~\ref{sec:exp}. Finally, Sec.~\ref{sec:conclusion} concludes the paper with a summary of contributions and outlines future directions.
 
\section{Methodology} \label{sec:method}

\begin{figure}
    \centering
    \includegraphics[width=0.9\linewidth]{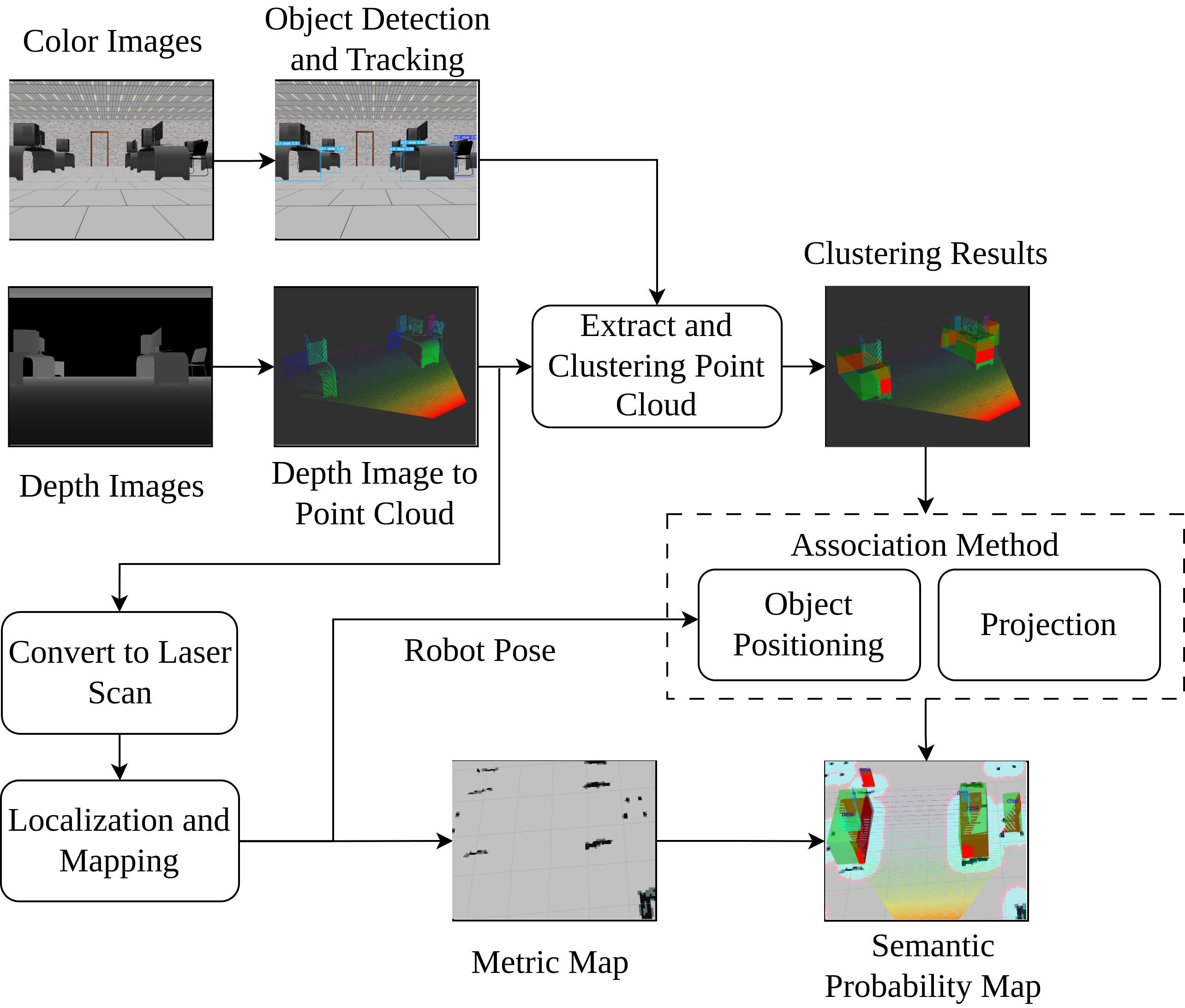}
    \caption{The overview of our proposal, front-end procedures: object detection, tracking, association; Back-end procedures: local mapping, loop closing, and global optimization based on CartoGrapher}
    \label{fig:overview}
\end{figure}

An overview of our proposed system is illustrated in Fig.~\ref{fig:overview}. Normally, the robot employs Simultaneous Localization and Mapping (SLAM) to construct occupancy grid maps. Our primary goal is to harness the RGB-D images to construct a semantic probability map based on these grid maps. First of all, the RGB images undergo object detection through a neural network model. For this task, We have opted for the highly accurate and real-time YOLOv8 architecture. Then, the detected objects are tracked using the BoT-SORT~\cite{aharon2022bot} algorithm to ensure consistent identification across different time frames. Second, we converted depth images to point clouds and extracted point cloud information of detected objects. To enhance data quality, a clustering method is employed to remove outlier point clouds. The core of semantic map creation involves the CartoGrapher method~\cite{hess2016real}, which tracks the robot's pose in the environment and generates a 2D metric map using 2D scan data derived from the point cloud. Subsequently, we calculate object positioning and project corresponding semantic information onto the robot's coordinate system. Finally, we enrich the metric map by incorporating semantic information and associated projection data, employing a probabilistic approach.

\subsection{Semantic knowledge understanding}
\begin{figure*}[!ht]
    \centering
    \begin{subfigure}[b]{0.28\textwidth}
    \centering
    \includegraphics[width=\textwidth]{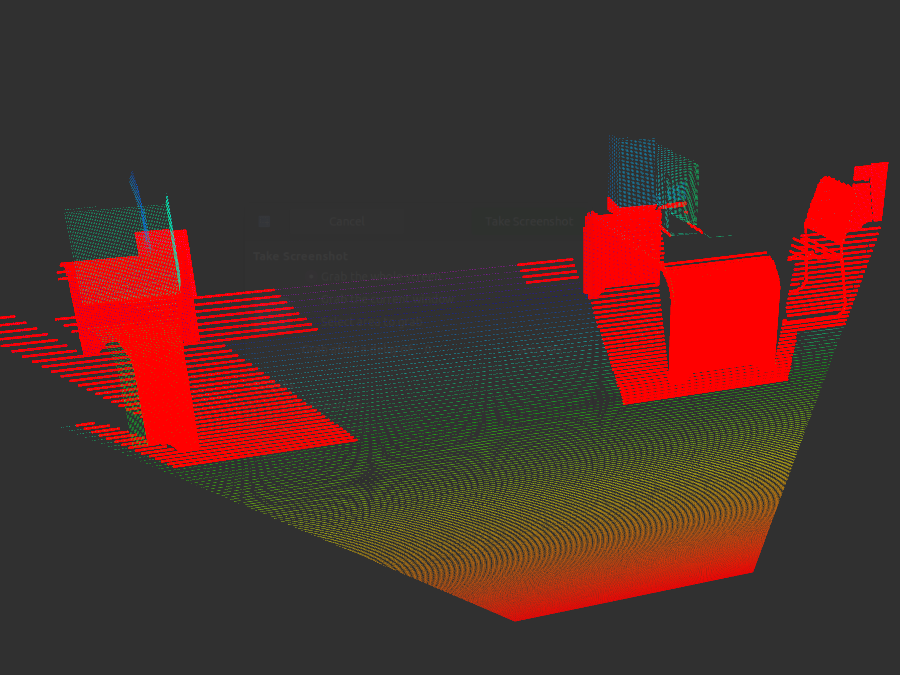}
    \caption{Raw point cloud of object class}
    \label{fig:tumrpetrans}
    \end{subfigure}
    \centering
    \begin{subfigure}[b]{0.28\textwidth}
    \centering
    \includegraphics[width=\textwidth]{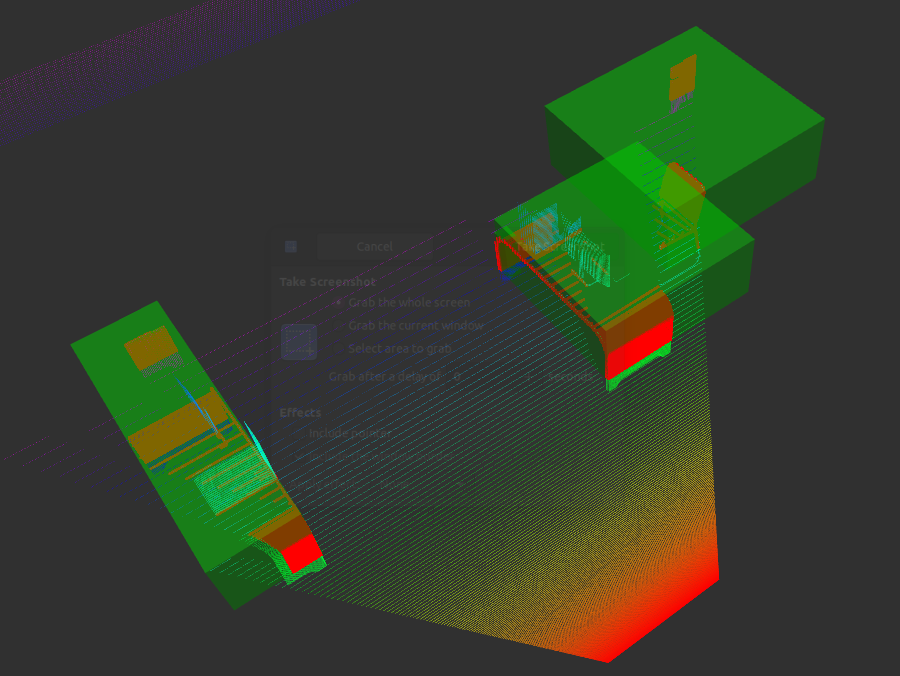}
    \caption{Remove background step}
    \label{fig:tumrperot}
    \end{subfigure}
    \centering
    \begin{subfigure}[b]{0.28\textwidth}
    \centering
    \includegraphics[width=\textwidth]{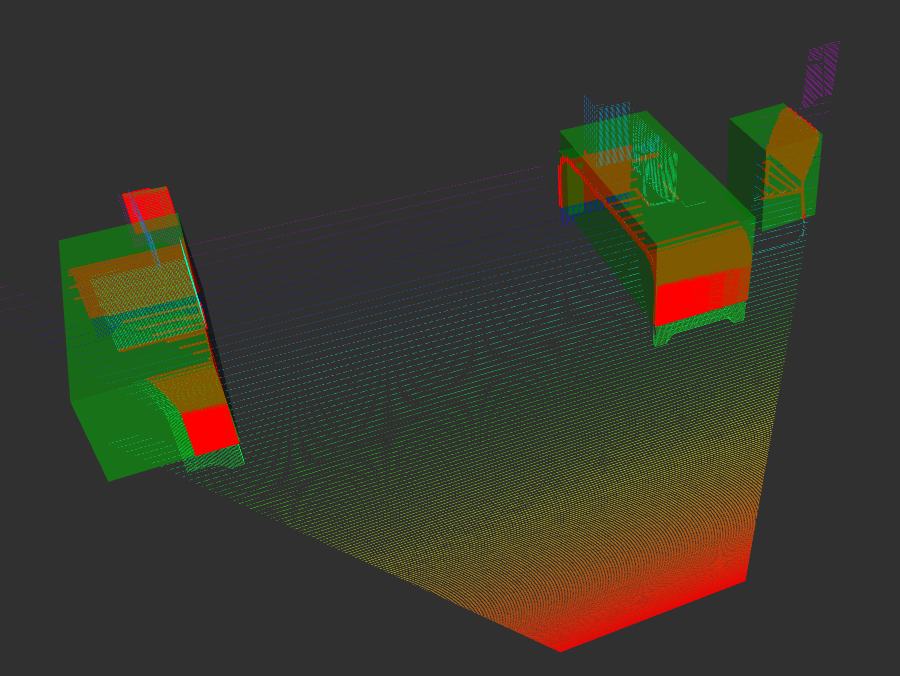}
    \caption{Cluster point cloud step}
    \label{fig:gazeborperot}
    \end{subfigure}
    
    \caption{Processing steps and results for semantic knowledge understanding. Red Point is a point cloud extracted from RGB images, and the green box is a 3D bounding box marker.}
    \label{fig:pre-process}
\end{figure*}
The detection of object classes within the field-of-view (FOV) of the RGB-D camera is a crucial step in our system. For that, we leverage a neural network designed to extract pixel-wise semantic information from the images. In our experiments, we employ the "You only look once"~\cite{redmon2016you} (Yolo) algorithm,  which stands as the state-of-the-art for object detection and real-time capabilities even when running on low-cost embedded devices. Specifically, we have opted for Yolov8 - the fastest, lightest with the highest precision-recall scores version. YOLOv8 takes the RGB original image as input and proceeds to generate bounding boxes and probabilities for each detected object class. Each bounding box is characterized by four parameters: center position $x(i), y(i)$, width $w(i)$, and height $h(i)$. Following object class detection, we implement BoT-SORT~\cite{aharon2022bot} to track multiple objects across various observations. BoT-SORT comprises three key components: a discrete Kalman Filter to model the object’s motion in the image plane, Camera Motion Compensation to compensate for the rigid camera motion, and IoU-Re-ID fusion to integrate appearance features into the tracker. The objects model of the class $i$ are defined as:

\begin{equation}
\textbf{x}_i = \begin{bmatrix}
x(i) & y(i) & w(i) & h(i) & \dot{x(i)} & \dot{y(i)} & \dot{w(i)} & \dot{h(i)}
\end{bmatrix}^T
\end{equation}
\begin{equation}
\textbf{z}_i = \begin{bmatrix}
z_x(i) & z_y(i) & z_w(i) & z_h(i)
\end{bmatrix}^T
\end{equation}

On the other hand, we extract point clouds captured by the RGB-D camera. Let's denote the depth value at pixel $(x, y)$ as $D(x, y)$. Assuming that the RGB-D camera is calibrated as a pin-hole camera model with the focal length $(f_x, f_y)$ and the optical center $(c_x, c_y)$, the intrinsic camera matrix $\mathbf{K} \in \mathbb{R}^{3\times3}$, and the extrinsic parameters $(\textbf{R}\in \mathbb{R}^{3\times3}, \textbf{T}\in \mathbb{R}^{3\times1})$. The 3D coordinate of the point $k$ in the world coordinate system represents:
\begin{equation}
    \textbf{p}_k = \begin{bmatrix}
X_k\\ 
Y_k\\ 
Z_K \\
1
\end{bmatrix} = \begin{bmatrix}
\textbf{R} & \textbf{T}\\ 
0_{0 \times 3} & 1
\end{bmatrix} * \begin{bmatrix}
D(x,y) K^{-1}x\\ 
D(x,y) K^{-1}y\\ 
D(x,y)\\
1
\end{bmatrix}
\end{equation}

In the subsequent phase, we proceed to extract complete point clouds $\mathbf{P}$ within the bounding box of the object class, which is obtained from the object detection model. To ensure robustness and efficiency, point cloud $\mathbf{P}$ undergoes a pre-processing stage, which includes two critical steps: Euclidean-based clustering and background removal. Euclidean-based clustering relies on the principle of spatial proximity. Points that exist within a specified Euclidean distance threshold $\epsilon$ are grouped into the same cluster, as they are likely to belong to the same object or surface. The Euclidean distance $d_{euclidean}$ between any two points $(x_1, y_1, z_1)$ and $(x_2, y_2, z_2)$ is computed as:
\begin{equation}
    d_{euclidean} = \sqrt{(x_1 - x_2)^2 + (y_1 - y_2)^2 + (z_1 - z_2)^2}
\end{equation}
Once the point cloud is segmented into clusters, we select the largest cluster based on the number of points it contains. Fig.~\ref{fig:pre-process} shows the processing step of the point cloud processing.
 
\subsection{Localization and mapping}
In pursuit of robust and precise localization, as well as comprehensive mapping capabilities, our research employs CartoGrapher, a cutting-edge 2D SLAM solution, seamlessly integrated within the Robot Operating System (ROS). This strategic combination empowers our robotic system to engage in simultaneous localization and mapping, thereby creating intricate representations of the surrounding environment. CartoGrapher represents a formidable 2D SLAM algorithm renowned for its ability to generate highly accurate maps of both indoor and outdoor environments. Key facets of CartoGrapher encompass:
\begin{itemize}
    \item Scan matching: CartoGrapher employs sophisticated scan-matching techniques to meticulously align consecutive laser scans. This alignment minimizes pose estimation errors, thereby enhancing the precision of the constructed map.
    \item Loop closure detection: This feature identifies previously visited locations within the environment, facilitating the correction of accumulated localization errors and ensuring map consistency.
\end{itemize}

The output is a 2D metric map representing the environment based on the grid $\mathbf{M}$ and robot's pose $\mathbf{p}_r \in \mathbb{R}^{3}$ in the map:
\begin{equation}
    \mathbf{p}_r =^M\textbf{M}_r = \begin{bmatrix}
        x & y & \theta
    \end{bmatrix}^T
\end{equation}
where $(x, y)$ and $\theta$ are the position and the orientation respectively, and $^M\textbf{M}_r$ is transformation matrix between map coordinate and robot coordinate.

\subsection{Semantic association}
After extracting the point cloud of the object class, we find the object positioning in map coordinates. We denote $\textbf{p}_o \in 
\mathbb{R}^{3 \times3}$ is the object position in the camera frame, which is as follows:
\begin{equation}
     \mathbf{p}_o =^c\textbf{M}_o = mean (\textbf{P}_o)
\end{equation}
where $\textbf{P}_o$ is the point cloud set of the object class.

Let $^r\textbf{T}_c$, $^r\textbf{R}_c$ are the transformation matrix, translation matrix, and rotation matrix between robot coordinates and camera coordinates, respectively. The object position in the map frame $\mathbf{p}_m = ^M\textbf{M}_o \in \mathbb{R}^{3}$ is shown as:

\begin{equation}
\mathbf{p}_m =
    \begin{bmatrix}
         ^M\textbf{M}_o \\ 1
    \end{bmatrix} = \begin{bmatrix}
        ^M\textbf{M}_r \\ 1
    \end{bmatrix} + ^r\textbf{M}_c \times \begin{bmatrix}
        ^c\textbf{M}_o \\ 1
    \end{bmatrix}
\end{equation}
in which, $^r\textbf{M}_c = \begin{bmatrix}
    ^r\textbf{R}_c & ^r\textbf{T}_c\\ 
0_{0 \times 3} & 1
\end{bmatrix}$ is the transformation matrix between robot coordinates and camera coordinates

For reliable navigation,  it is imperative that robots effectively navigate around obstacles in their environment. However, conventional mapping systems can often lead robots to collide when it comes to objects with hollow bottoms, as they are typically represented as free grid spaces in metric maps. To address this challenge, we employ the RANdom-SAmple Consensus (RANSAC)~\cite{fischler1981random} for the projection method. The core principle of RANSAC is to iteratively sample subsets of data points, hypothesize models, and evaluate the model's consensus with the data points. For instance, the plane $\mathbb{P}$ representing a "chair seat" object has three degrees of freedom:
\begin{equation}
    \mathbb{P}: \textbf{n}^T \textbf{p}_m + D = 0
\end{equation}
where $\textbf{n} \in \mathbb{R}^{3}$ is the coefficients of the plane's normal vector and $d$ represents the distance from the origin to the plane along the normal vector, effectively determining the plane's offset.
\begin{figure}[!ht]
    \centering
    \begin{subfigure}[b]{0.24\textwidth}
    \centering
    \includegraphics[width=\textwidth]{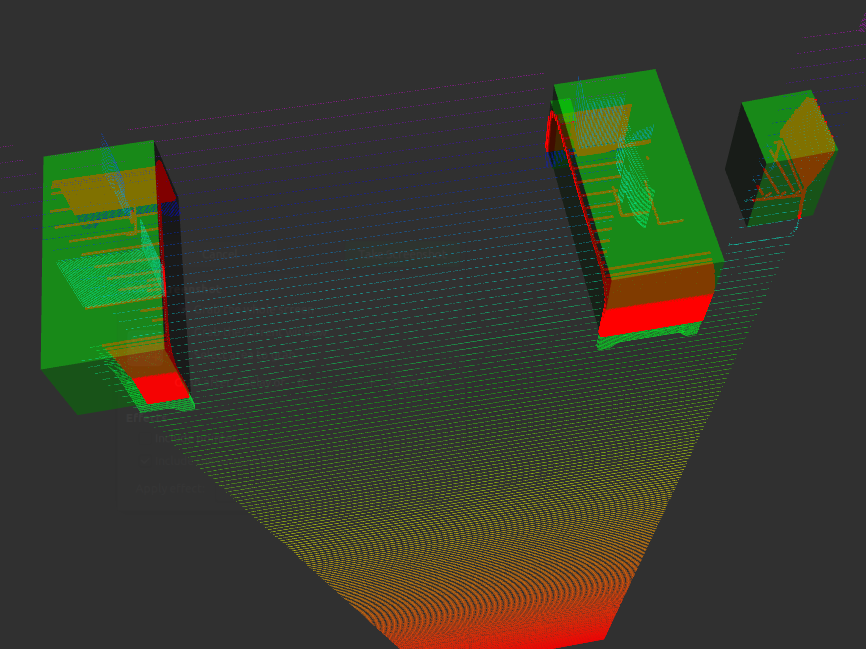}
    \caption{Before projection}
    \end{subfigure}
    \centering
    \begin{subfigure}[b]{0.24\textwidth}
    \centering
    \includegraphics[width=\textwidth]{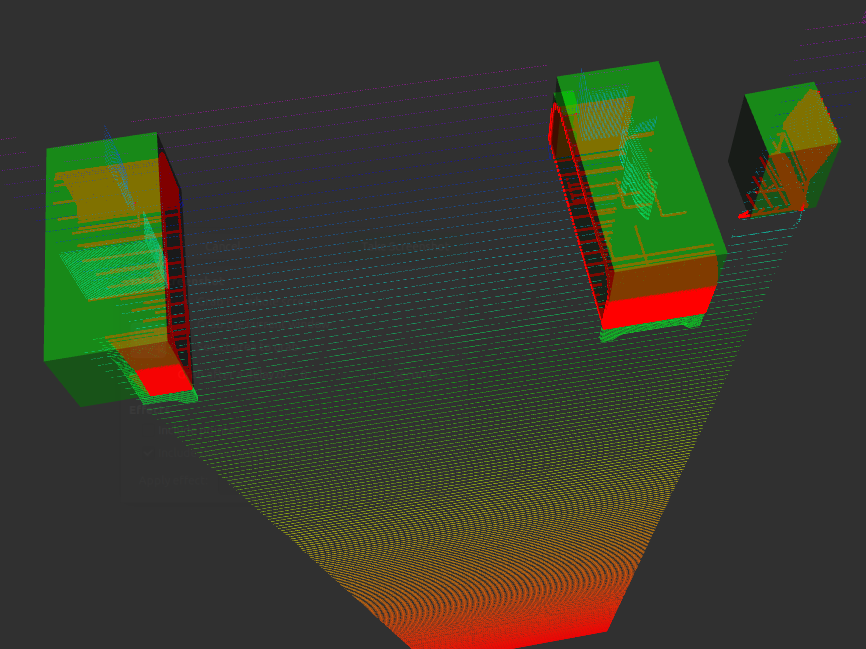}
    \caption{After projection}
    \end{subfigure}
    
    \caption{Projection of the 3D detected object point clouds}
    \label{fig:project}
\end{figure}

After this segment plane $\mathbb{P}$ using RANSAC, each point $\textbf{p}_k$ in this plane is represented in the camera frame with $z$ pointing upward. Its projection to the map plane thus can be conducted by simply setting its $z$ coordinate of the detected plane to coincide with the map plane. Fig. \ref{fig:project} describes the projected process.

\subsection{Probabilistic map representation}
\begin{figure}[!ht]
    \centering
    \includegraphics[width=0.5\linewidth]{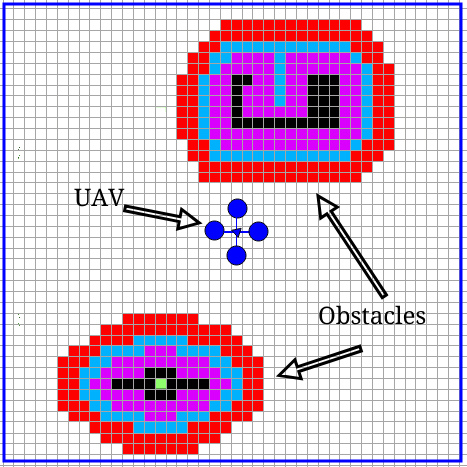}
    \caption{Probabilistic map representation}
    \label{fig:probabilistic_map}
\end{figure}
In the pursuit of enhancing scene comprehension and decision-making capabilities in real-world environments, we introduce a probabilistic semantic representation framework that seamlessly integrates with the 2D costmap. The 2D costmap serves as a foundational grid-based representation of the environment, encoding essential navigation information such as obstacle locations and traversal costs. Fig.~\ref{fig:probabilistic_map} shows an example of a grid costmap where each cell is assigned a value ranging from 0 to 255 corresponding to the probability of 0\% to 100\% of the cell being occupied. Building upon this, our system injects a layer of semantic richness into the costmap by associating each grid cell with probabilistic semantic attributes. Instead of merely categorizing cells as obstacles or free space, we introduce probabilities representing the likelihood of specific object classes being present within each cell. This dynamic representation enables our autonomous system to not only perceive the spatial distribution of objects but also gauge the uncertainty associated with each detection. By amalgamating costmap information with probabilistic semantics, our system is empowered to make informed navigation decisions that factor in the likelihood of encountering specific objects, thus enhancing safety and adaptability in complex and dynamically changing environments. This approach enriches the conventional 2D costmap with semantic intelligence, unlocking new avenues for reliable and context-aware navigation.
\section{Results} \label{sec:exp}
\subsection{System Setup and Dataset}
\begin{figure}
    \centering
    \includegraphics[width=0.6\linewidth]{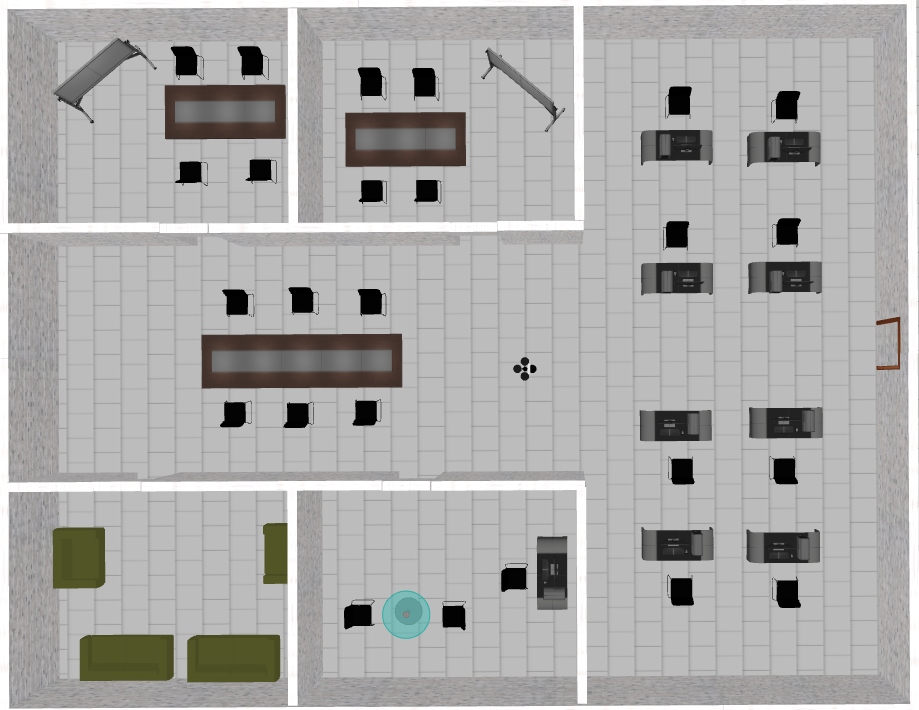}
    \caption{UAV and environments simulation}
    \label{fig:env}
\end{figure}

The experimental evaluation of our proposed system as illustrated in Fig.~\ref{fig:env} was executed using the Hummingbird UAV platform, which is equipped with a RealSense camera. The Hummingbird UAV distinguishes itself with its lightweight design, facilitating agile flight maneuvers and accurate navigation in complex and dynamic environments, such as those encountered in demanding search and rescue missions. The RealSense D455 camera complements the UAV’s capabilities by providing RGB-D data, which has a horizontal FOV is 90 deg, vertical FOV  is 58, depth FOV is 98, image size is $640 \times 480$, and frame per second (FPS) is 60 Hz.

In the context of this research, the training images have 6 types of objects: chair, coffee table, conference table, sofa, whiteboard, and desk. The chair, the desk, the coffee table, and the whiteboard are fully hollow bottom objects; the conference table is a party hollow bottom object and the sofa is a non-hollow bottom object. To this end, we trained the YOLOv8 object detector with 300+ different images. 
\subsection{Experimental Results}
\begin{figure}[!ht]
    \centering
    \begin{subfigure}[b]{0.24\textwidth}
    \centering
    \includegraphics[width=\textwidth]{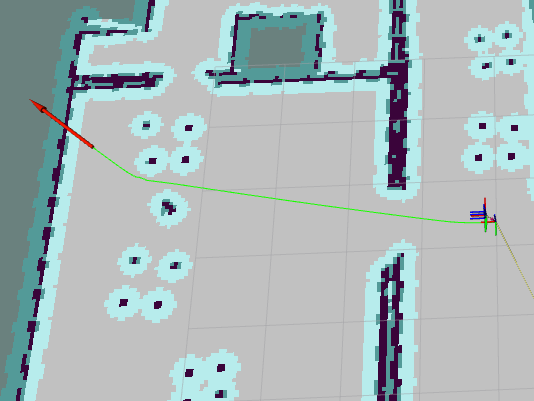}
    \caption{Obstacle avoidance path using only metric map}
    \end{subfigure}
    \centering
    \begin{subfigure}[b]{0.24\textwidth}
    \centering
    \includegraphics[width=\textwidth]{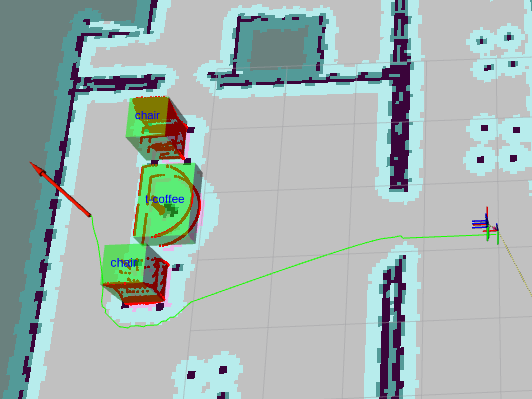}
    \caption{Obstacle avoidance path using only our map}
    \end{subfigure}
    
    \caption{Avoiding coffee table obstacles}
    \label{fig:table}
\end{figure}

\begin{table}[ht]
  \centering
  \caption{Results of detection models}
  \label{tab:obj}
  \begin{tabular}{p{0.08
\textwidth}|p{0.12\textwidth}|p{0.065\textwidth}|p{0.065\textwidth}|p{0.04\textwidth}}
    \toprule
    \textbf{Models} & \textbf{mAP (IOU=0.5)
} & \textbf{Parameter} & \textbf{Precision} & \textbf{Recall} \\
    \midrule
    Yolov3~\cite{bersan2018semantic}&90.0\% & \hspace{0.05cm} 8.7M & 85.9\% & 84.6\%\\
    Fast R-CNN & 95.2\% & 12.9M & 90.2\% & 92.0\%\\
    MobileNet & 88.6\% &\hspace{0.05cm} 4.6M & 91.1\% & 83.7\%\\
    Yolov4 & 94.8\% & 60.0M & 82.6\%& 86.4\%\\
    RTMDet & 95.9\% & 52.3M & 91.5\% & 88.4\%\\  
    Yolov5 & 94.0\% & \hspace{0.05cm} 7.0M & 87.0\% & 92.7\% \\
    Yolov7 & 94.8\% & \hspace{0.05cm} \textbf{3.7}M & 83.8\% & 96.2\% \\
    Our & \textbf{98.2}\% & 11.1M & 92.0\%& 92.9\%\\
    \bottomrule
  \end{tabular}
\end{table}

\begin{figure*}[!ht] 
    \centering
    \begin{subfigure}[b]{0.24\textwidth}
    \centering
    \includegraphics[width=\textwidth]{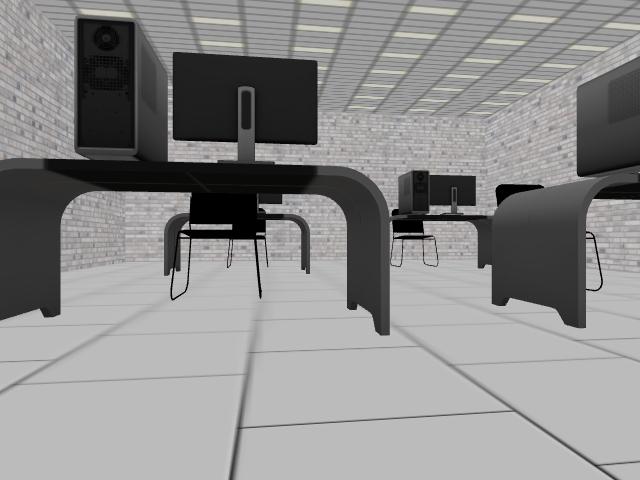}
    \end{subfigure}
    \begin{subfigure}[b]{0.24\textwidth} 
    \centering
    \includegraphics[width=\textwidth]{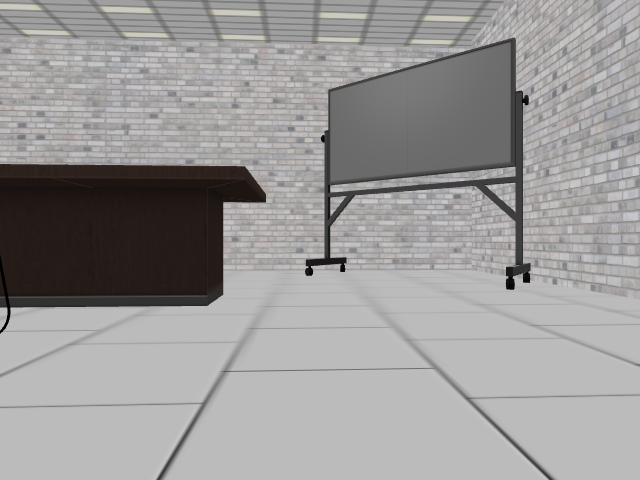}
    \end{subfigure}
    \begin{subfigure}[b]{0.24\textwidth}
    \centering
    \includegraphics[width=\textwidth]{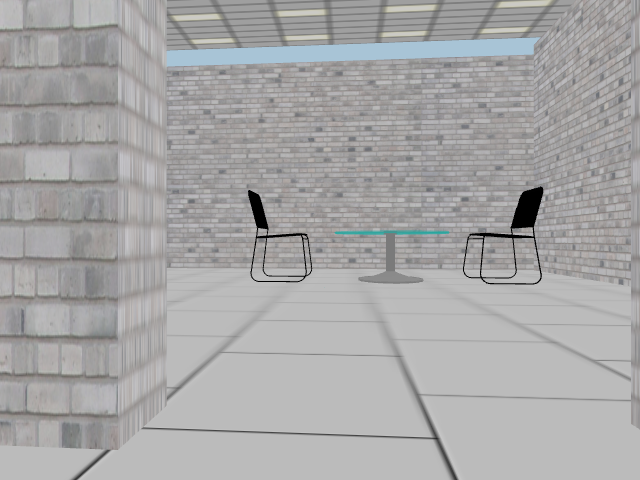}
    \end{subfigure}
    \begin{subfigure}[b]{0.24\textwidth}
    \centering
    \includegraphics[width=\textwidth]{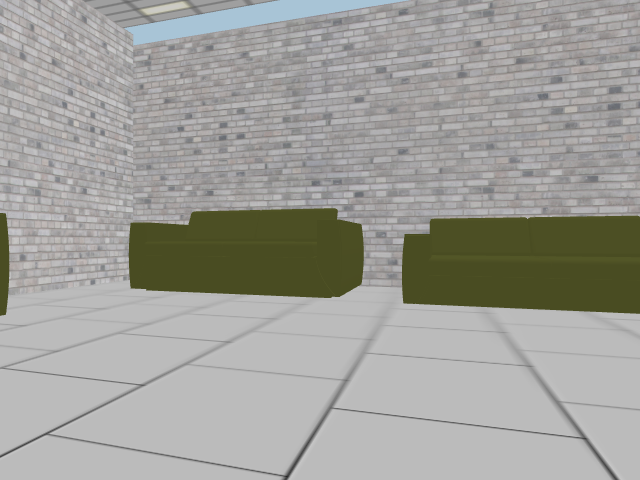}
    \end{subfigure}
    \caption*{(a) Input image from camera}

        \centering
    \begin{subfigure}[b]{0.24\textwidth}
    \centering
    \includegraphics[width=\textwidth]{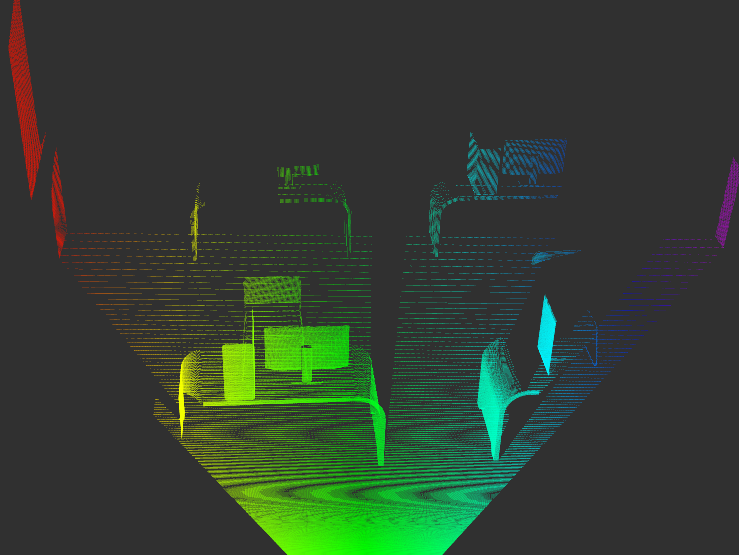}
    \end{subfigure}
    \begin{subfigure}[b]{0.24\textwidth} 
    \centering
    \includegraphics[width=\textwidth]{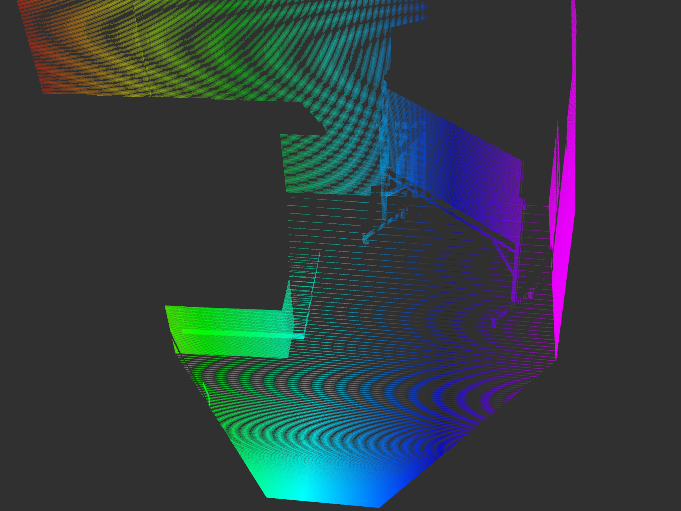}
    \end{subfigure}
    \begin{subfigure}[b]{0.24\textwidth}
    \centering
    \includegraphics[width=\textwidth]{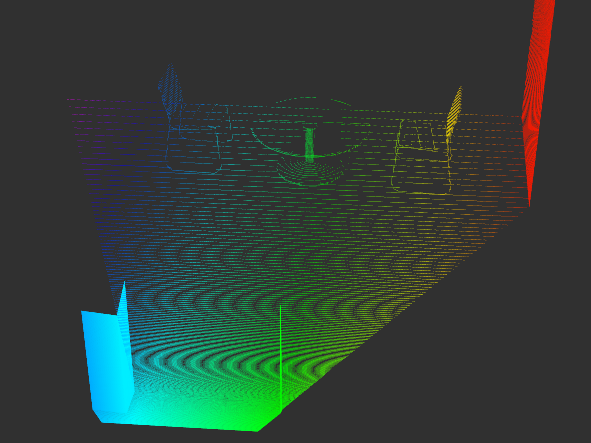}
    \end{subfigure}
    \begin{subfigure}[b]{0.24\textwidth}
    \centering
    \includegraphics[width=\textwidth]{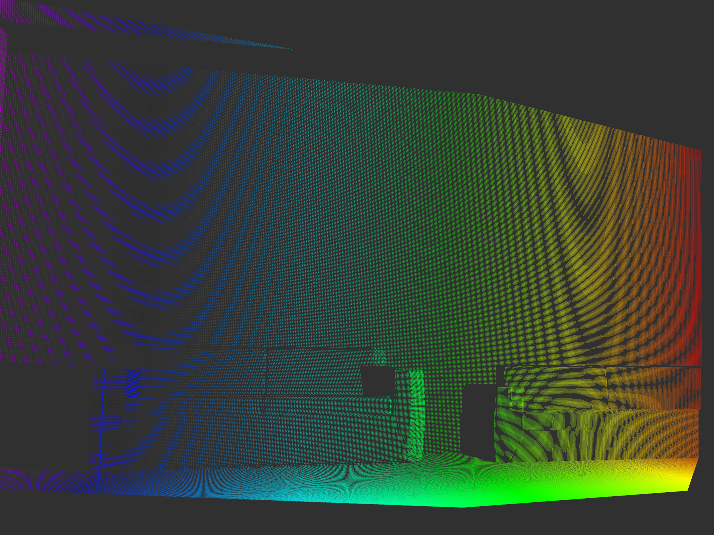}
    \end{subfigure}
    \caption*{(b) Point cloud from input images}
    
    \centering
    \begin{subfigure}[b]{0.24\textwidth}
    \centering
    \includegraphics[width=\textwidth]{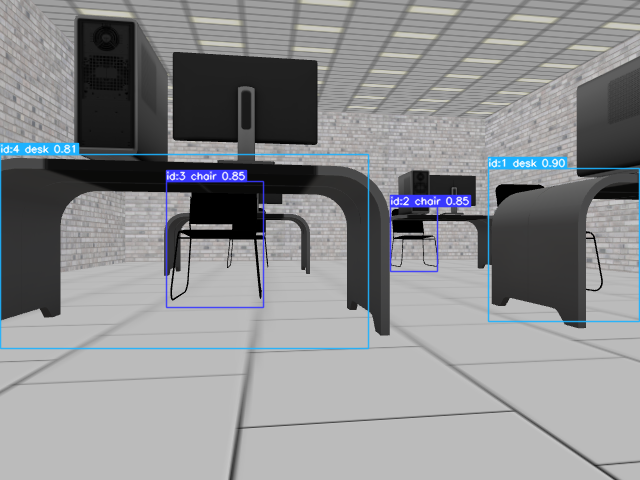}
    \end{subfigure}
    \begin{subfigure}[b]{0.24\textwidth} 
    \centering
    \includegraphics[width=\textwidth]{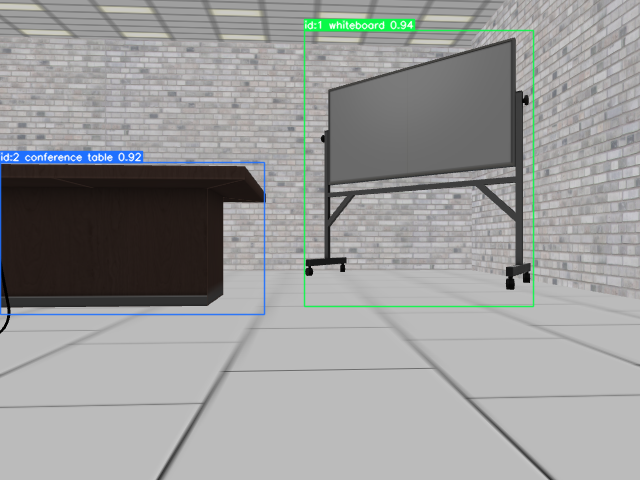}
    \end{subfigure}
    \begin{subfigure}[b]{0.24\textwidth}
    \centering
    \includegraphics[width=\textwidth]{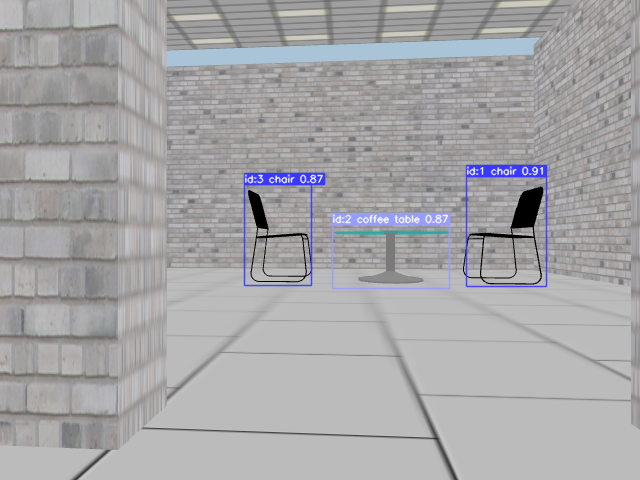}
    \end{subfigure}
    \begin{subfigure}[b]{0.24\textwidth}
    \centering
    \includegraphics[width=\textwidth]{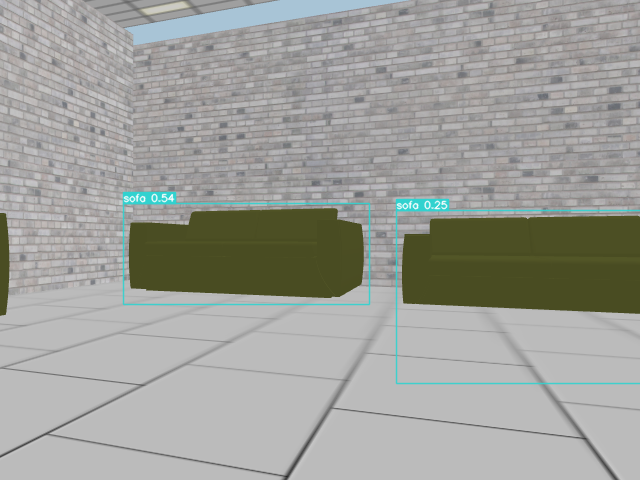}
    \end{subfigure}
    \caption*{(c) Object detection from input images}
    
    \centering
    \begin{subfigure}[b]{0.24\textwidth}
    \centering
    \includegraphics[width=\textwidth]{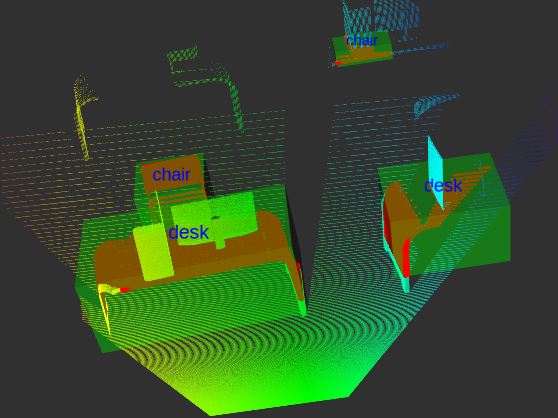}
    \end{subfigure}
    \begin{subfigure}[b]{0.24\textwidth} 
    \centering
    \includegraphics[width=\textwidth]{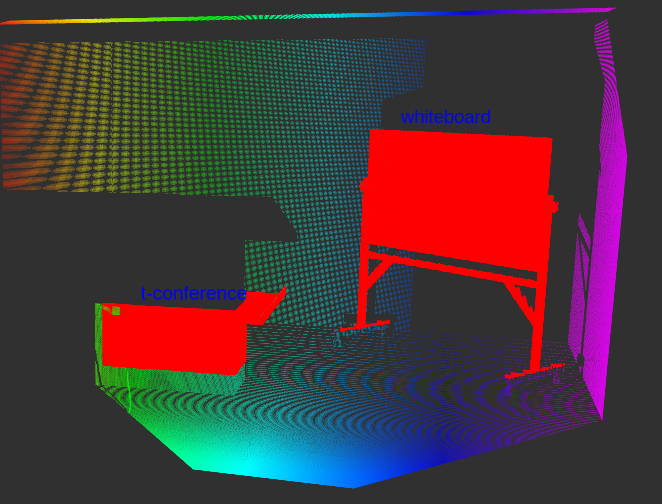}
    \end{subfigure}
    \begin{subfigure}[b]{0.24\textwidth}
    \centering
    \includegraphics[width=\textwidth]{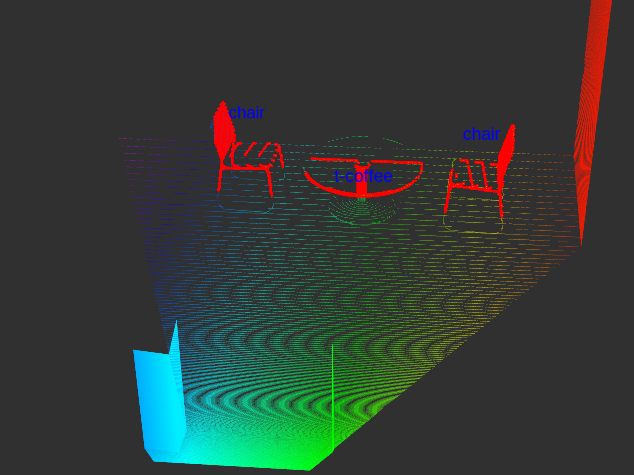}
    \end{subfigure}
    \begin{subfigure}[b]{0.24\textwidth}
    \centering
    \includegraphics[width=\textwidth]{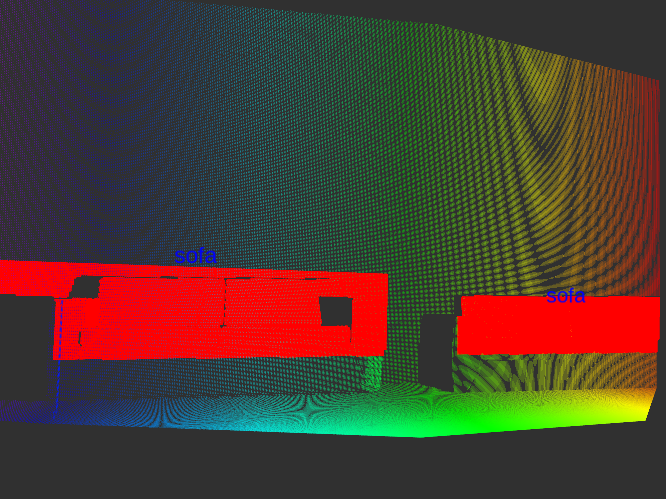}
    \end{subfigure}
    \caption*{(d) Point cloud clustering}

    \centering
    \begin{subfigure}[b]{0.24\textwidth}
    \centering
    \includegraphics[width=\textwidth]{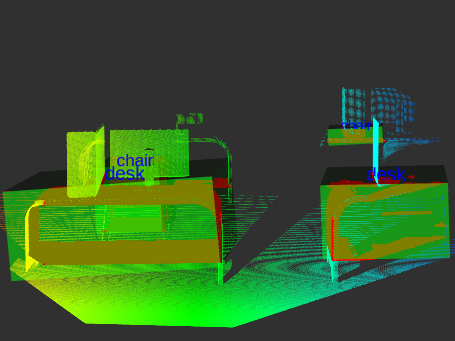}
    \end{subfigure}
    \begin{subfigure}[b]{0.24\textwidth} 
    \centering
    \includegraphics[width=\textwidth]{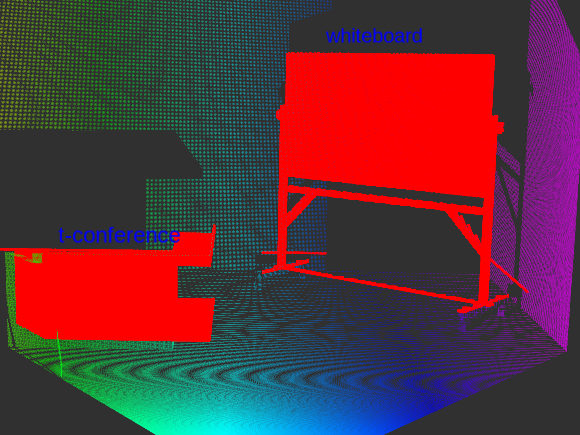}
    \end{subfigure}
    \begin{subfigure}[b]{0.24\textwidth}
    \centering
    \includegraphics[width=\textwidth]{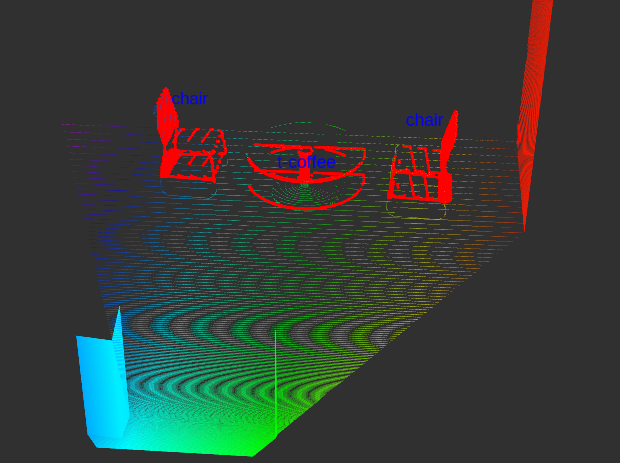}
    \end{subfigure}
    \begin{subfigure}[b]{0.24\textwidth}
    \centering
    \includegraphics[width=\textwidth]{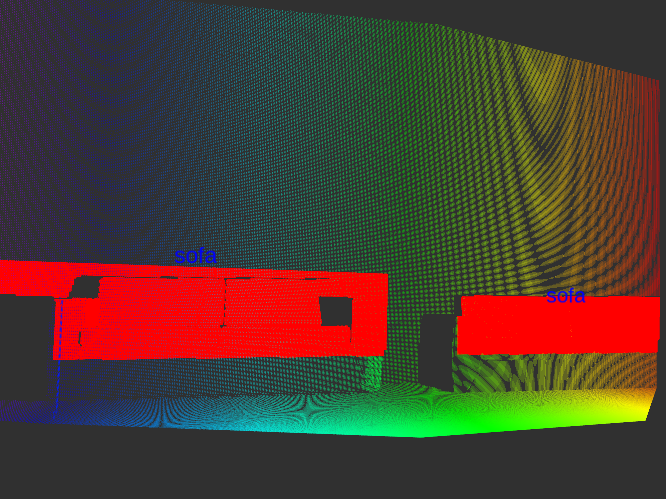}
    \end{subfigure}
    \caption*{(e) Point cloud projection}

    \centering
    \begin{subfigure}[b]{0.24\textwidth}
    \centering
    \includegraphics[width=\textwidth]{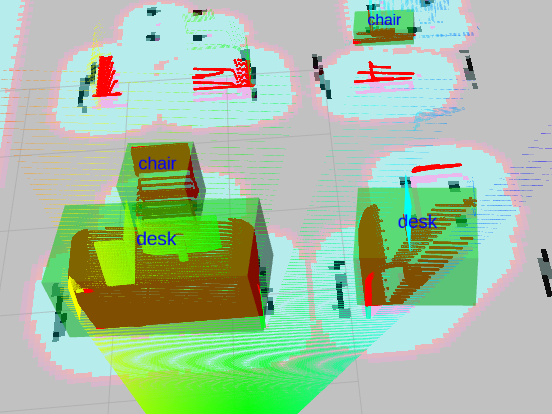}
    \end{subfigure}
    \begin{subfigure}[b]{0.24\textwidth} 
    \centering
    \includegraphics[width=\textwidth]{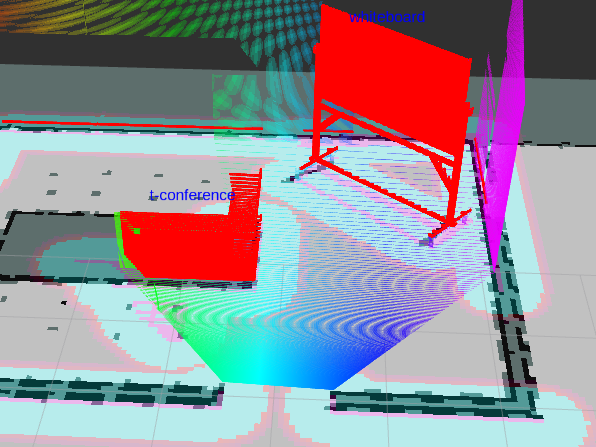}
    \end{subfigure}
    \begin{subfigure}[b]{0.24\textwidth}
    \centering
    \includegraphics[width=\textwidth]{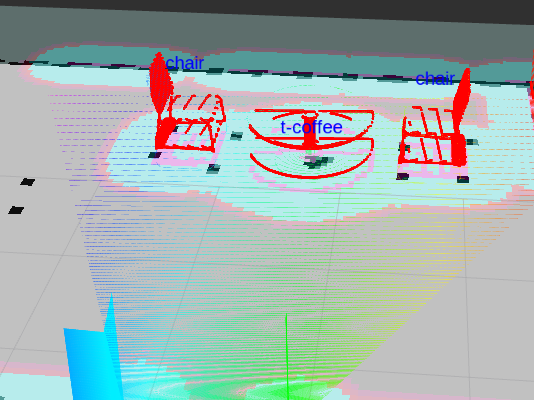}
    \end{subfigure}
    \begin{subfigure}[b]{0.24\textwidth}
    \centering
    \includegraphics[width=\textwidth]{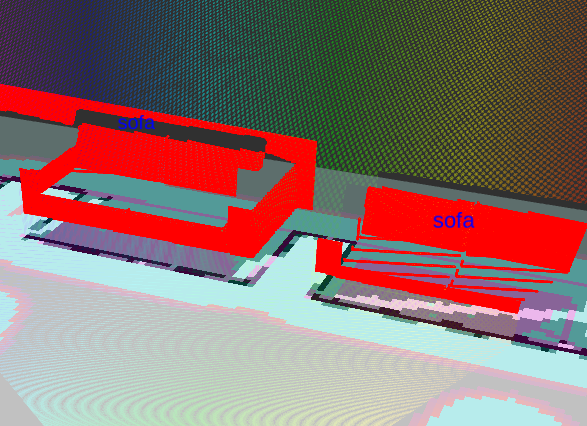}
    \end{subfigure}
    \caption*{(f) Probabilistic map representation}
    \caption{Visual representation of the obtained semantic maps}
    \label{fig:semantic_map}
\end{figure*}

\textit{1) Semantic Mapping:} Fig.~\ref{fig:semantic_map} visually illustrates experimental results for the visual representation of the obtained semantic maps where red points are point clouds of the detected object, and blue text is the object's label. Our semantic map includes object information, object position and project it in metric map. Areas around the object are occupancy probability in each cell, which helps the UAVs keep a safe distance from the object. 
To demonstrate the efficiency of the object detection model, we conducted a comparative evaluation with several prominent object detection models, as outlined in Table. \ref{tab:obj}. Notably, we evaluate our methodology with Yolov3, which was employed in a prior work by D. Bersan~\cite{bersan2018semantic}. Our results reveal that our system achieves exceptional accuracy, surpassing 98\%. This high accuracy enables the detection of multiple objects, even when they overlap. Besides that, a slight latency in the point cloud processing stage occasionally led to objects being associated with incorrect locations in the 2D map, particularly when the robot executed turns. To address this issue, we implemented synchronized object detection and point cloud processing stages. Furthermore, the experimental results show that the projection stage demonstrates the ability to determine the pose of the object, and the probabilistic map representation stage can enhance the map information in each cell compared to just 0 for free cells and 1 for occupied cells of the regular metric map.

\begin{figure}[!ht]
    \centering
    \begin{subfigure}[b]{0.24\textwidth}
    \centering
    \includegraphics[width=\textwidth]{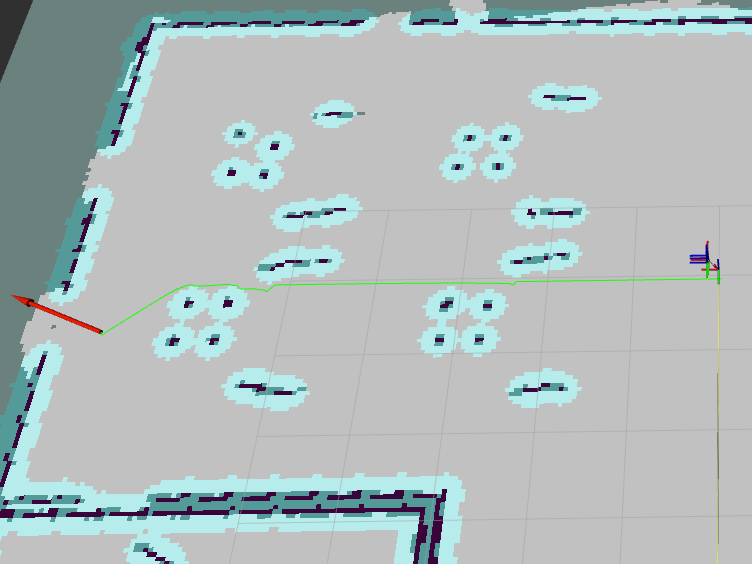}
    \caption{Obstacle avoidance path using only metric map}
    \end{subfigure}
    \centering
    \begin{subfigure}[b]{0.24\textwidth}
    \centering
    \includegraphics[width=\textwidth]{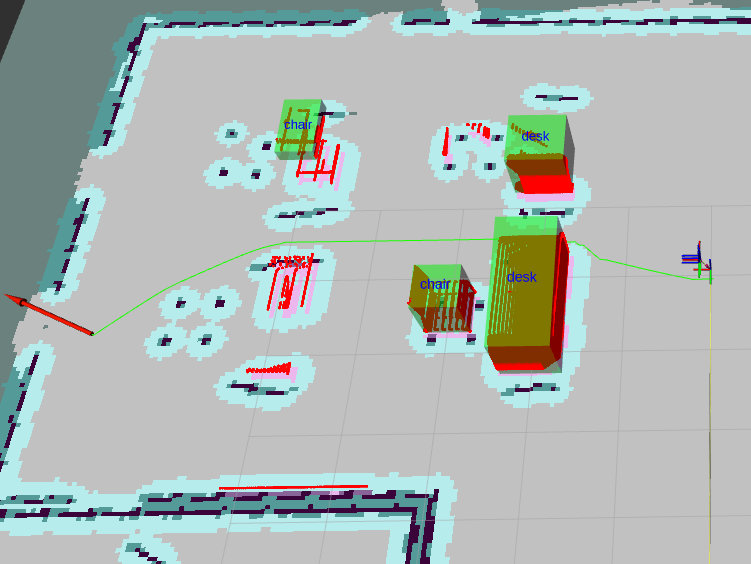}
    \caption{Obstacle avoidance path using only our map}
    \end{subfigure}
    
    \caption{Avoiding chair and desk obstacles}
    \label{fig:chair}
\end{figure}

\textit{2) Safety Navigation:} To demonstrate the ability of safety navigation, we test our methodology with chair obstacles, desk obstacles, and coffee table obstacles, which are fully hollow bottom objects. Fig.~\ref{fig:chair} and Fig.~\ref{fig:table} show the trajectory of the UAV when avoiding the obstacles. When relying solely on metric maps for obstacle avoidance, navigating through the hollow spaces of these objects can expose the UAV to collision risks. In contrast, our methodology possesses the ability to precisely identify object positions and sizes. This knowledge empowers the UAV to navigate strategically and safely, avoiding collisions even in intricate environments. In addition, the object detection stage and the point cloud process stage consume only 0.25s and 0.5s, respectively. This real-time performance underscores the practical operability of our approach, making it well-suited for time-sensitive applications such as navigation.

\section{Conclusions} \label{sec:conclusion}
In this paper, we proposed a methodology for autonomous navigation, particularly in high-stakes scenarios like search and rescue operations. By merging object detection, and 2D SLAM, we've created a probabilistic semantic map. This map enhances traditional metric maps by incorporating object-specific attributes, such as object class, precise positioning, and probabilistic information. Our experiments demonstrate the system's ability to generate semantic maps, even recognize complex objects like chairs and desks, and safely navigate through hollow bottom objects. The practical deployment on an embedded computer validates its real-world utility. In summary, our work significantly advances autonomous robotics, promising safer and more efficient operations in diverse domains. Future research can further enhance semantic understanding and navigation robustness.

\section*{Acknowledgment}
This work was supported by the Asian Office of Aerospace Research and Development under Grant/Cooperative Agreement Award No. FA2386-22-1-4042.
\bibliographystyle{IEEEtran}
\bibliography{ref}
\vspace{12pt}
\color{red}

\end{document}